\pdfoutput=1
\documentclass[10pt,twocolumn,letterpaper]{article}

\usepackage{cvpr}
\usepackage{times}
\usepackage{graphicx}
\usepackage{amsmath}
\usepackage{amssymb}
\usepackage{enumitem}

\usepackage{wasysym}
\usepackage[nice]{nicefrac}       
\usepackage{algorithm}
\usepackage{algorithmic}

\newcommand{\Nystrom}{Nystr\"{o}m}

\newcommand{\op}[1]{\ensuremath{\mathcal{#1}}}

\newcommand{\mat}[1]{\ensuremath{\mathbf{#1}}}
\newcommand{\matsub}[2]{\ensuremath{\mathbf{#1}_{#2}}}

\newcommand{\matcol}[2]{\ensuremath{\mathbf{#1}^{(#2)}}}
\newcommand{\matelem}[3]{\ensuremath{\mathbf{#1}_{#2,#3}}}
\newcommand{\mata}[1]{\ensuremath{\mathbf{\tilde{#1}}}}
\newcommand{\matasub}[2]{\ensuremath{\mathbf{\tilde{#1}}_{#2}}}
\newcommand{\matamet}[2]{\ensuremath{\mathbf{\tilde{#1}}^{#2}}}

\newcommand{\ve}[1]{\ensuremath{\mathbf{#1}}}

\newcommand{\ves}[2]{\ensuremath{\mathbf{#1_{#2}}}}

\newcommand{\real}[1]{\ensuremath{\mathbb{R}^{{#1}}}}

\newcommand{\On}[1]{\ensuremath{\mathcal{O} \left({#1}\right)}}

\newcommand{\normzero}[1]{\ensuremath{\|{#1}\|_0}}

\newcommand{\normtwo}[1]{\ensuremath{\|{#1}\|_2}}
\newcommand{\normfro}[1]{\ensuremath{\|{#1}\|_F}}

\newcommand{\dataset}[1]{\textit{#1}}

\graphicspath{{./}}
\DeclareGraphicsExtensions{.pdf,.png}

\usepackage[bookmarks=false]{hyperref}

\cvprfinalcopy 

\def\cvprPaperID{3870} 

\ifcvprfinal\fi
\begin{document}

\title{Efficient, sparse representation of manifold distance matrices for classical scaling}

\author{Javier S. Turek\\
Intel Labs, Hillsboro, Oregon\\
{\tt\small javier.turek@intel.com}
\and
Alexander G. Huth\\
Departments of Computer Science \& Neuroscience, The University of Texas at Austin\\
{\tt\small huth@cs.utexas.edu}
}

\maketitle

\begin{abstract}
Geodesic distance matrices can reveal shape properties that are largely invariant to non-rigid deformations, and thus are often used to analyze and represent 3-D shapes. However, these matrices grow quadratically with the number of points. Thus for large point sets it is common to use a low-rank approximation to the distance matrix, which fits in memory and can be efficiently analyzed using methods such as multidimensional scaling (MDS). In this paper we present a novel sparse method for efficiently representing geodesic distance matrices using biharmonic interpolation. This method exploits knowledge of the data manifold to learn a sparse interpolation operator that approximates distances using a subset of points. We show that our method is 2x faster and uses 20x less memory than current leading methods for solving MDS on large point sets, with similar quality. This enables analyses of large point sets that were previously infeasible.
\end{abstract}

\section{Introduction}
\label{sec:introduction}

\begin{figure}
	\begin{center}
		\includegraphics[width=\columnwidth]{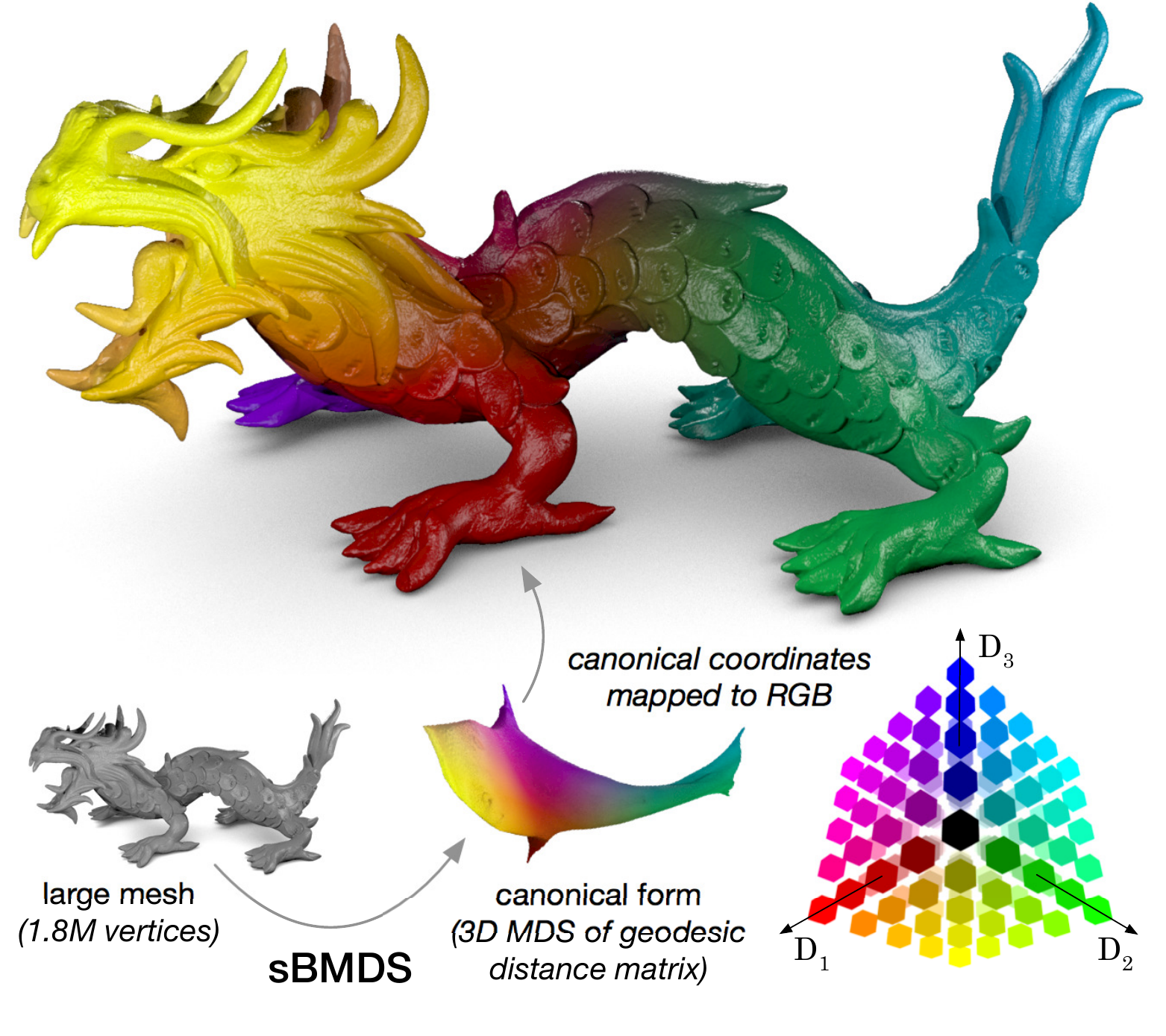} 
    		\caption{\emph{Sparse Biharmonic Multidimensional Scaling} (sBMDS) was used to obtain the canonical form for a mesh with 1.8M vertices. MDS is impossible on the full distance matrix (1.8M x 1.8M, 26 TB), but easy using a sparse approximation (50,000 landmarks, 20.9 GB). Here the 3-D canonical coordinates found by MDS are are mapped into RGB colors on the original mesh.}
		\label{fig:dragon_mds}
	\end{center}
\end{figure}

Distance matrices have many important roles in machine learning, graphics, and computer vision. Yet with growing data, it is becoming impossible to store or process full distance matrices. For $n$ points, \On{n^2} space is required to store the distance matrix, and, depending on the type of data and distance metric, as much as \On{n^2\log n} time can be required to compute it. Many algorithms that operate on distance matrices execute in \On{n^3} time. With growing $n$ these requirements quickly become intractable.

One solution is to work with a low-rank approximation of the distance matrix. While the best rank $k$ approximation is given by the Singular Value Decomposition (SVD), its \On{n^3} computation is impractical for large matrices.
An alternative is the \Nystrom{} method \cite{nystrom1930}, which computes a low-rank approximation by subsampling $l$ columns from the distance matrix. 
\Nystrom{} has been studied theoretically \cite{NYSTROM_THEORY,NYSTROM_SAMPLING} and empirically \cite{NYSTROM_ACCURACY,NYSTROM_LARGE,NYSTROM_MANIFOLDLEARNING}, but it is a generic method that does not take advantage of any knowledge about the geometry of the point set.

For the specific problem of approximating geodesic distance matrices computed from 3-D meshes, methods such as spectral MDS (SMDS) \cite{SMDS} and fast MDS (FMDS) \cite{FMDS} have been proposed.
These methods were designed to compress large geodesic distance matrices so that they can be analyzed using multidimensional scaling (MDS). The key insight offered by SMDS and FMDS that is not exploited by methods such as \Nystrom{} is that the geometry of the distance matrix closely mirrors the geometry of the underlying point set.
If the point set lies along some manifold, then rows of the distance matrix will lie along a higher-dimensional projection of that same manifold.
Intuitively this follows from the fact that nearby points on a manifold will have similar distances to other points in the manifold.

Here we explore a novel method, biharmonic matrix approximation (BHA), and its sparse form sBHA.
Like SMDS and FMDS, our method uses biharmonic interpolation to approximate the full distance matrix from a few samples.
However, we improve upon those methods by noting that most of the values in the biharmonic interpolation operator are very close to zero, and thus the operator can be represented sparsely with little to no effect on accuracy of the approximation.
Sparsification also increases the efficiency of algorithms that use the approximate distance matrix, such as MDS.
Our method thus makes it possible to approximate and operate on extremely large point sets where existing methods would suffer poor performance due to memory constraints (Figure \ref{fig:dragon_mds}).

\section{Background}

\label{sec:background}
Given a set of $n$ points from a manifold $\op{M}$ embedded in $\real{d}$, $\mathcal{X} = \left\{ \ves{x}{i} \right\}_{i=1}^n$ with $\ves{x}{i} \in \op{M}$, we define the geodesic distance matrix $\mat{K}\in \real{n \times n}$ as
\begin{equation}
\matelem{K}{i}{j} = d_\op{M}\left(\ves{x}{i},\ves{x}{j}\right),
\label{eq:KernelDefinition}
\end{equation}
where \matelem{K}{i}{j} represents the element in row $i$ and column $j$ of the  matrix \mat{K}, and $d_\op{M}(\cdot,\cdot)$ is the geodesic distance, or the length of the shortest path between two points along the surface of $\op{M}$. Assume that $d_\op{M}(\cdot,\cdot)$ is symmetric for a pair of points, i.e., $d_\op{M}(\ves{x}{i},\ves{x}{j})=d_\op{M}(\ves{x}{j},\ves{x}{i})$.

\subsection{Biharmonic interpolation}
Let $g$ be a real-valued function defined on a smooth manifold $\op{M}$ embedded in $\real{d}$. The manifold has an associated Laplace-Beltrami Operator (LBO), $\Delta$, that depends on the Riemannian metric of the manifold \cite{Reuter2009} such that $\Delta g = \mbox{div }( \mbox{grad } g )$.

Now suppose that we are given $g(\ves{b}{i})$ for a set of $l$ points, $\left\{ \ves{b}{i} \right\}_{i=1}^l \in \mathcal{M}$, and wish to interpolate $g(\ves{u}{i})$ at a different set of $m$ points, $\left\{ \ves{u}{i} \right\}_{i=1}^m \in \mathcal{M}$. One solution is biharmonic interpolation, which finds a smooth function (i.e. one with continuous second derivative) that passes exactly through $g(\ves{b}{i})$ for all $i$. Biharmonic interpolation is accomplished by finding a solution to the biharmonic equation $\Delta^2 g(\ve{u}) = 0$, subject to Dirichlet boundary conditions given by $g(\ve{b})$. Note that this is equivalent to exactly minimizing the smoothness-preserving energy function defined in \cite{SMDS} and \cite{FMDS}.

In the discrete case the biharmonic operator $\Delta^2$ is specified as a sparse matrix \mat{M}. We assume that the manifold consists exclusively of the data points $\ve{b}$ and $\ve{u}$, and thus that \mat{M} is an $(l+m) \times (l+m) = n \times n$ matrix.
We can organize $\mat{M}$ so that the first $l$ rows and columns correspond to $\ve{b}$, our known data points, and the last $m$ rows and columns correspond to $\ve{u}$, the points at which we wish to interpolate $g$. Thus \mat{M} can be split into four submatrices,
\begin{equation}
  \mat{M} = \begin{bmatrix} \mat{M}_{bb} & \mat{M}_{bu} \\ \mat{M}_{ub} & \mat{M}_{uu} \end{bmatrix}.
  \label{eq:BiharmonicOperator}
\end{equation}
To find the interpolated values we then solve for $\hat{g}(\ve{u})$ in the modified biharmonic equation
\begin{eqnarray*}
  \begin{bmatrix} \mat{M}_{bb} & \mat{M}_{bu} \\ \mat{M}_{ub} & \mat{M}_{uu} \end{bmatrix}\begin{bmatrix} g(\ve{b}) \\  \hat{g}(\ve{u}) \end{bmatrix} &=& \ve{0},\\
  \mat{M}_{ub} g(\ve{b}) + \mat{M}_{uu} \hat{g}(\ve{u}) &=& \ve{0},
\end{eqnarray*}
yielding the solution
\begin{eqnarray}
\hat{g}(\ve{u}) &=& -\mat{M}_{uu}^{-1} \mat{M}_{ub} g(\ve{b}),
\label{eq:SystemLinEquations}
\end{eqnarray}
which is a fully specified, sparse linear system of equations that can be solved using standard methods. Note that $\hat{g}(\ve{u})$ is related to $g(\ve{b})$ through a linear transformation that is independent of the values in $g(\ve{b})$. We can think of this linear transformation, $-\mat{M}_{uu}^{-1}\mat{M}_{ub}$, as an \emph{interpolation operator}: a transformation that will interpolate any function known on the points $\ve{b}$ onto the points $\ve{u}$.

\subsection{Obtaining the discrete biharmonic operator}
\label{sec:biharmonic_operator}
The discrete biharmonic operator $\mat{M}$ is given as \cite{Lipman2010}
\begin{equation*}
\label{eq:bho}
\mat{M} = (\mat{V} - \mat{A})^\top \mat{D}^{-1} (\mat{V} - \mat{A}),
\end{equation*}
where $\mat{D}$ is a diagonal matrix containing the ``lumped mass" at each point, $\mat{A}$ is a weighted adjacency matrix, and $\mat{V}$ is a diagonal matrix containing the sum of the adjacency weights for each datapoint $\mat{V}_{i,i} = \sum_j \mat{A}_{i,j}$ \cite{Reuter2009}.


In some cases there are closed-form solutions for the lumped mass and weighted adjacency matrices. If the point set $\mathcal{X}$ is a triangular mesh embedded in 3-D space, the lumped mass $\mat{D}_{i,i}$ is $\nicefrac{1}{3}$ of the total area of the triangles incident on point $i$, and the adjacency weight
\begin{equation*}
\label{eq:cotadj}
\mat{A}_{i,j} = \frac{\cot(\alpha_{i,j}) + \cot(\beta_{i,j})}{2},
\end{equation*}
where $\alpha_{i,j}$ and $\beta_{i,j}$ are the angles opposite the edge between points $i$ and $j$ \cite{Reuter2009,crane2013}. Similar solutions for $\mat{D}$ and $\mat{A}$ exist for point clouds \cite{Liu2012} and quad meshes \cite{Desbrun2008} in 3-D space. If $\mathcal{X}$ is a generic graph, we set $\mat{D}=\mat{I}$, and set \mat{A} equal to the graph adjacency matrix, where $\matsub{A}{i,j}=1$ if nodes $i$ and $j$ share an edge, and $0$ otherwise.

If no graph is given, \mat{D} and \mat{A} can be estimated. Here it is common to again set $\mat{D}=\mat{I}$ and then estimate $\mat{A}$ using some weighted nearest neighbor algorithm. We will not address this further here, but note that the weights given by Stochastic Neighbor Embedding (SNE) \cite{Hinton2002} perform particularly well at this problem.

\subsection{Multidimensional Scaling}
Given a matrix \mat{K} with distances between all pairs of $n$ points, Multidimensional Scaling (MDS) methods compute a low-dimensional embedding of these $n$ points. The embedded coordinates $\left\{ \ves{z}{i} \right\}_{i=1}^n \in \real{m}$ are found by minimizing, for all pairs of points, the difference between their Euclidean distances in the embedding \normtwo{\ves{z}{i} - \ves{z}{j}} and their squared distance $\matelem{K}{i}{j}^2$ in the original space:
\begin{equation}
	\arg \min_{\mat{Z}} \normfro{\mat{Z}\mat{Z}^T + \frac{1}{2}\mat{J}\mat{E}\mat{J}},
	\label{eq:MDS_problem}
\end{equation}
where $\mat{Z} = \left[\ves{z}{1},\dots,\ves{z}{n}\right]^T \in \real{n \times m}$ contains the embedded coordinates, $\matelem{E}{i}{j}= \matelem{K}{i}{j}^2$ are the squared distances, and $\mat{J} =  \mat{I} - \nicefrac{1}{n}\mat{1}\mat{1}^T$ is a centering matrix with \mat{1} being a column vector of ones.

In classical MDS the optimal solution to Problem \eqref{eq:MDS_problem} is obtained by first computing the eigenvalue decomposition $\mat{V}\mat{\Lambda}\mat{V}^T$ of the matrix $-\nicefrac{1}{2} \mat{J}\mat{E}\mat{J}$, and then truncating its decomposition to the biggest $m$ eigenvalues, $\mat{\Lambda}_{m}$, and their respective eigenvectors $\mat{V}_{m}$. The embedded points are then computed as $\mat{Z}= \mat{V}_{m}\mat{\Lambda}_{m}^{1/2}$.

Solving Problem \eqref{eq:MDS_problem} requires storing the matrix \mat{E} in memory, which is prohibitive for more than a few tens of thousands of points.
To overcome this limitation, alternative methods use a low-rank approximation of \mat{E}. This is achieved by subsampling the points and interpolating their distances. Methods such as Landmark MDS (LMDS) \cite{LMDS}, SMDS \cite{SMDS}, and FMDS \cite{FMDS} propose different solutions to this approximation problem. In what follows, we present an alternative method to approximate \mat{E} with a low-rank and {\em sparse} approximation, simultaneously enabling a smaller memory footprint and faster MDS algorithm for very large numbers of points.

\section{Biharmonic matrix approximation (BHA)}

\label{sec:methods}
In the proposed method we use biharmonic interpolation to approximate a manifold-structured distance matrix $\mat{K}$. We exploit the fact that the manifold structure of $\mat{K}$ is similar to that of the underlying point set $\mathcal{X} = \{\ves{x}{i}\}_{i=1}^n$. We assume that $\mathcal{X}$ lives in a manifold $\op{M}$ whose biharmonic operator $\mat{M}$ can either be computed directly or approximated as described in Section \ref{sec:biharmonic_operator}. 

The first step is to select $\ve{b}$, a set of $l$ landmark points from $\mathcal{X}$. Landmarks are selected using an iterative farthest point procedure \cite{FSP}: the first landmark $\ves{b}{1}$ is selected at random from $\mathcal{X}$, and the geodesic distance from that point to all the other points, $\mat{K}_{\ves{b}{1},\cdot}$ is computed. 
Each subsequent landmark is then chosen as the point with the largest minimum distance to any of the current landmarks,
\begin{equation}
  \ves{b}{j} = \underset{\ves{x}{i}}{\mbox{argmax }}\left(\mbox{min }_{t=1}^{j-1} \mat{K}_{\ves{b}{t},\ves{x}{i}}\right)
  \label{eq:FarthestPoint}
\end{equation}
until $l$ landmarks have been selected.

The second step in BHA is to create an interpolation operator $\mat{P}$ that interpolates values from the landmarks $\ve{b}$ to the entire manifold by solving Equation \ref{eq:SystemLinEquations},
\begin{equation}
\label{eq:PDefinition}
\mat{P} =
\begin{bmatrix}
\mat{I}_l \\
-\mat{M}_{uu}^{-1} \mat{M}_{ub}
\end{bmatrix} =
\begin{bmatrix}
\mat{I}_l \\
\matsub{P}{u}
\end{bmatrix},
\end{equation}
where $\{\ves{u}{i}\} = \mathcal{X} \setminus \{\ves{b}{i}\}$ is the set of all non-landmark points, and $\mat{M}_{uu}$ and $\mat{M}_{ub}$ are defined as in Equation \eqref{eq:BiharmonicOperator}.

The third step in BHA is to form $\mat{W} \in \real{l \times l}$, the diagonal block from $\mat{K}$ that contains geodesic distances between landmark points: $\matelem{W}{i}{j} = d_{\op{M}}(\ves{b}{i}, \ves{b}{j})$ for $1 \le i,j \le l$. Note that these values were already computed during the landmark selection procedure, and can be re-used here.

The \emph{Biharmonic Matrix Approximation (BHA)} is then obtained as
\begin{equation}
\label{eq:PWPt}
\matamet{K}{\mathit{BHA}} = \mat{P} \mat{W} \mat{P}^\top.
\end{equation}
BHA can be thought of as performing two interpolation operations. First, the product $\mat{P}\mat{W}$ interpolates geodesic distances from the landmarks to all the points, approximating the $n \times l$ submatrix formed by the columns of $\mat{K}$ corresponding to the landmarks. Right-multiplying by $\mat{P}^\top$ then interpolates each row of $\mat{P} \mat{W}$ across the entire manifold, giving the full $n \times n$ approximation $\matamet{K}{\mathit{BHA}}$.

\subsection{From dense to sparse}
\label{sec:sparsity}

\begin{figure}[t]
	\begin{center}
		\includegraphics[trim=20 20 40 20, width=0.93\linewidth]{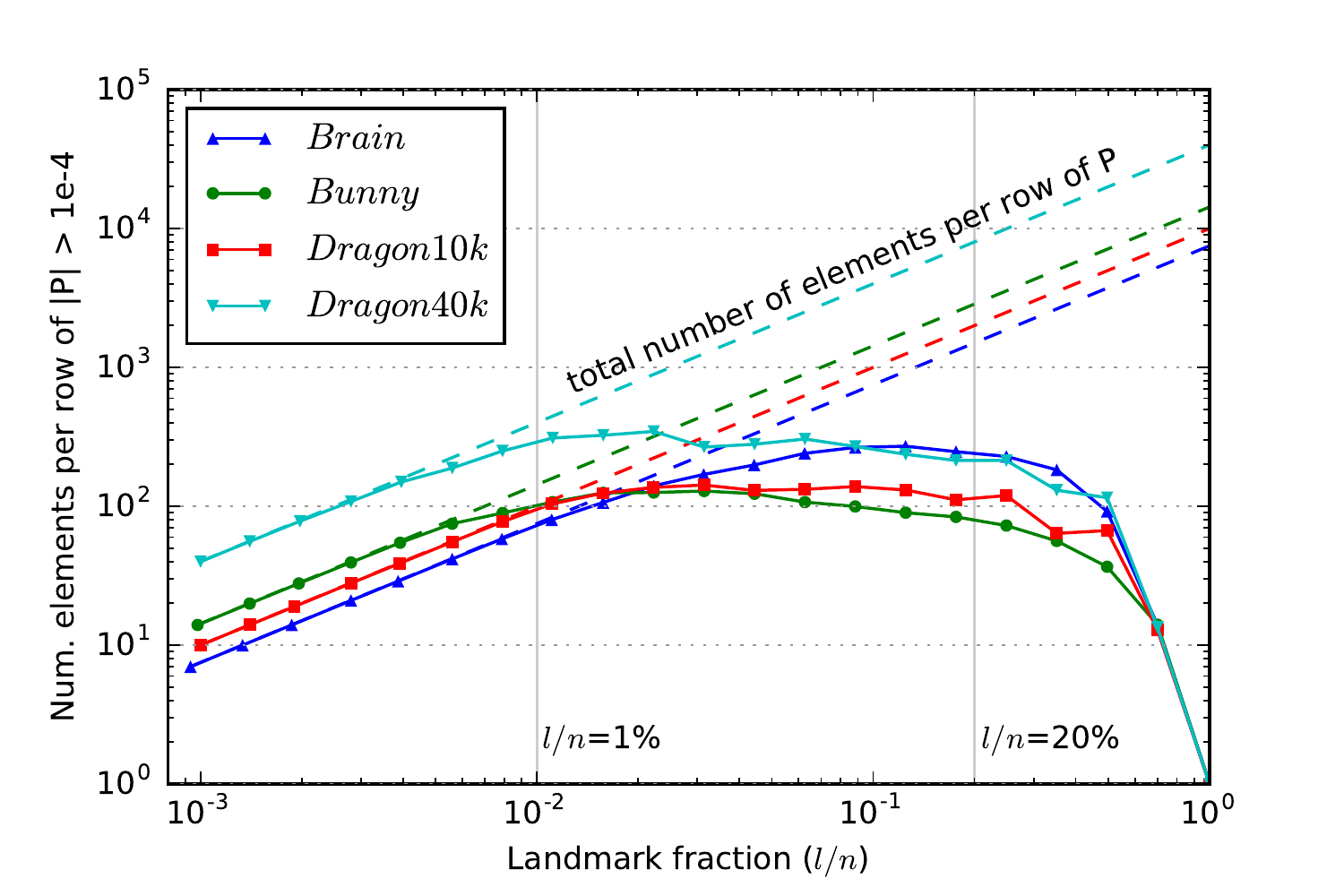}
	\end{center}
	\caption{Sparsity of \mat{P} as a function of number of landmarks. The number of large elements per row of \mat{P} peaks at landmark fractions $l/n < 20\%$ for all point sets.}
	\label{fig:Psparsity}
\end{figure}

Analyzing the interpolation operator \mat{P} reveals some useful numerical properties. The coefficients in \mat{P} define weights that are used to interpolate values in \mat{W} onto the non-landmark points $\ves{u}{i}$. Intuitively, as the number of landmarks grows, the number of data points influenced by each landmark will shrink. In the limit of the number of landmarks approaching the total number of data points, $\mat{P} \rightarrow I$, and each data point is only influenced by a single landmark. Thus, even though biharmonic interpolation operators are not theoretically guaranteed to be sparse (i.e. they do not have compact support), empirically most entries in $\mat{P}$ tend to be near zero (Figure \ref{fig:Psparsity}). 

Therefore, instead of computing the interpolation operator \mat{P} given in Equation  \eqref{eq:PDefinition}, we propose to compute a sparse interpolation operator. We note that Equation \eqref{eq:PDefinition} describes the solution to a system of linear equations to obtain \mat{P}. We replace the matrix inversion by a sparse coding problem of the form
\begin{eqnarray}
\label{eq:PSparseProblem}
	\matasub{P}{u} &=& \arg \min_{\matsub{P}{u}} \normfro{\mat{M}_{uu}\matsub{P}{u} + \mat{M}_{ub}}^2
 \quad \nonumber \\ &&\mathrm{s.t.}\quad \normzero{\matcol{P}{i}_{u}} \le p \quad \forall i,
\end{eqnarray}
where the \normzero{\ve{x}} is the $\ell_0$-quasinorm that measures the number of non-zeros in a vector \ve{x}, $\matsub{P}{u}$ is the submatrix of $\mat{P}$ corresponding to the non-landmark points $\{\ves{u}{i}\}$, $\matcol{P}{i}_{u}$ is the $i^{th}$ column of \matsub{P}{u}, and $p$ is a scalar value. Problem \eqref{eq:PSparseProblem} constrains the solution to be a sparse matrix \matasub{P}{u} with at most $p$ non-zeros per column. The sparse interpolation operator \matsub{P}{\mathit{sparse}} is obtained by plugging the solution \matasub{P}{u} into Equation \eqref{eq:PDefinition}. 
The \emph{Sparse Biharmonic Matrix Approximation (sBHA)} is then obtained as
\begin{equation}
\matamet{K}{\mathit{sBHA}} = \matsub{P}{\mathit{sparse}} \mat{W} \matsub{P}{\mathit{sparse}}^\top.
\end{equation}


Sparse coding is a combinatorial problem, so its solution is typically approximated using greedy methods or convex relaxations such as OMP \cite{OMP} or LASSO \cite{LASSO}. However, in cases such as this where most entries in \mat{P} are already very close to zero, we can use the much cheaper Thresholding method \cite{MIKIBOOK} to approximate Problem \eqref{eq:PSparseProblem}. Noting that the matrix $\mat{M}_{uu}$ is square and invertible, this method chooses the $p$ biggest elements in magnitude from the dense solution for each column of \mat{P}. Each column of \matsub{P}{\mathit{sparse}} is solved separately, requiring \On{n} memory to store one dense column of \mat{P} at a time, and \On{n} runtime complexity to find the biggest $p$ elements.
Typically, we are interested in $p_{\mathit{row}}$, the average number of non-zeros per row. The relation between $p$ and a $p_{\mathit{row}}$ parameter is given as $p = \nicefrac{(n-l) p_{\mathit{row}}}{l}$.


How big must $p_{\mathit{row}}$ be in order for \matsub{P}{\mathit{sparse}} to be a good approximation of \mat{P}? If a large $p_{\mathit{row}}$ is required, then there are no memory or runtime benefits of using \matsub{P}{\mathit{sparse}}. However, since most elements in \mat{P} are close to zero, it seems likely that $p_{\mathit{row}}$ does not need to be very large. Although we provide no theoretical proof, our empirical evaluation in Section \ref{sec:experiments} suggests that $p_{\mathit{row}}$ can be considered a small constant number. $p_{\mathit{row}} \le 50$ seems to work well for any acceptable number of landmarks for large point sets.

\subsection{Runtime and space complexity}
\label{sec:complexity}
To analyze the runtime and space complexity of BHA we divide the method into two steps: a preprocessing step that computes the biharmonic operator \mat{M}, and a step that computes the interpolation operator \mat{P} and the diagonal block matrix \mat{W} for the landmarks.

In the preprocessing step, we compute the lumped mass matrix \mat{D}, the adjacency matrix \mat{A} and the sum of adjacency weights \mat{V} in order to obtain the biharmonic operator \mat{M}. Matrices \mat{D} and \mat{V} are diagonal and can be computed with one pass over the adjacency matrix \mat{A}, which is also sparse. 
When the manifold \op{M} is given as a mesh with at most $t$ neighbors for each point, the adjacency matrix is given and there is no need for more computation or space than is required by these matrices, yielding a total runtime complexity \On{n t^2} and space \On{n t^2} to compute and store \mat{M}. Practically, $t$ has a small value up to tens of neighbors per data point, such that still $t^2 \ll n$.

In the low-rank approximation step we draw a set of landmark points and then compute the interpolation operator \mat{P} and the matrix \mat{W}.
Finding landmarks using the farthest point procedure requires \On{lf} time, where $f$ is the complexity of evaluating one row of the distance matrix \mat{K}.
Computing \mat{P} requires solving the system of linear equations described in \eqref{eq:SystemLinEquations}. There are $l$ right hand side vectors and the system has $n-l$ equations. As \mat{M} is a sparse matrix with \On{n} non-zeros (assuming $t \ll n$), Equation  \eqref{eq:SystemLinEquations} can be solved with sparse linear solvers in \On{T_{P}nl}, where $T_P$ is the number of iterations needed to converge to an specific accuracy. Alternatively, we can compute the sparse Cholesky decomposition of \matsub{M}{uu} and then solve for the $l$ right hand side vectors, having a runtime complexity \On{n^{1.5} + nl}.
In the dense case, computing \mat{P} requires \On{nl} space. In the sparse case, using $p_{\mathit{row}}$ non-zeros per row with $p_{\mathit{row}} < l$ and practically constant, the space is reduced to \On{np_{\mathit{row}}} non-zeros. Computing sparse columns adds no cost. Computing the submatrix \mat{W} of \mat{K} requires \On{l^2} space and \On{lf} time, but \mat{W} can re-use the distances computed while selecting landmarks.
Thus sBHA has runtime complexity of \On{n t^2 + T_{P}nl + lf} and space complexity of \On{\max\{np_{\mathit{row}}+l^2, nt^2\}}. 

\subsection{Classical Scaling with BHA}
After computing the low-rank approximation \mata{E} of the matrix \mat{E} of squared distances, one can solve Classical Scaling by extracting the eigenvalues of $-\nicefrac{1}{2}\mat{J}\mata{E}\mat{J}= -\nicefrac{1}{2}\mat{J}\mat{P}\mat{W}\mat{P}^T\mat{J}$.
We remind the reader that our aim is to compute the $m$ biggest eigenvalues and respective eigenvectors without computing  a prohibitive  $n \times n$ matrix. 
When using BHA, we follow the method proposed by \cite{FMDS}. We dub this implementation \emph{Biharmonic Multidimensional Scaling (BMDS)}. First we compute the QR decomposition  $\mat{Q}\mat{R}=\mat{J}\mat{P}$, where $\mat{R} \in \real{l \times l}$.
Then, we compute the eigen-decomposition of the matrix $-\nicefrac{1}{2} \mat{R}\mat{W}\mat{R}^T$ given by $\mata{V}{\Lambda}\mata{V}^T$. The embedding is computed as $\mat{Z}=\mat{Q}\mata{V}_{m}\mat{\Lambda}_{m}^{1/2}$. Overall runtime complexity for this method is \On{nl^2 +l^3}.

The above solution computes a QR decomposition that requires \On{nl} memory to store matrix \mat{Q}. This solution works, but its memory usage can be prohibitive when $n$ is very large.
Alternatively, we propose \emph{sparse BMDS (sBMDS)}, which uses the Lanczos method \cite{parlett1998symmetric} to compute only the needed $m$ eigenvalues and eigenvectors. The Lanczos method requires only matrix-vector multiplications, avoiding the storage of big matrices \cite{parlett1998symmetric}.
This multiplication is very fast as the interpolation matrix $\mat{P}_{sparse}$ is sparse, the matrix \mat{J} is a sum of identity and a rank-one matrix, and \mat{W} is small compared to $n$. Overall, computing the $m$ eigenvalues and eigenvectors with such matrix-vector multiplications has a runtime complexity of \On{mn+mnp_{\mathit{row}}+ml^2+ m^2} with a small additional space for a vector of length \On{n}.

\subsection{Relationships to other approximation methods}

\subsubsection{\Nystrom{}}
\label{sec:nystrom}
The \Nystrom{} method \cite{nystrom1930} is a symmetric matrix approximation that has been successfully applied to machine learning problems on many datasets.
The method requires one to sample a subset of $l$ landmark points from the point set and compute the submatrix $\mat{C}\in \real{n \times l}$, which consists of the corresponding $l$ columns of \mat{K}.
The low-rank approximation is then obtained as
\begin{equation}
\matamet{K}{\mathit{Nys}} = \mat{C} \mat{W}^{\dagger}\mat{C}^\top,
\label{eq:NystromApprox}
\end{equation}
where $\mat{W}\in\real{l \times l}$ is the symmetric diagonal block corresponding to the columns and rows of \mat{K} for the landmarks, and $\mat{W}^{\dagger}$ is its Moore-Penrose pseudoinverse.
While BHA may appear structurally similar to \Nystrom{}, a critical difference is that BHA does not compute the pseudoinverse $\mat{W}^\dagger$. This renders BHA more stable than \Nystrom{} in situations where \mat{W} is (near-)singular.
The largest difference between BHA and \Nystrom{} is that \Nystrom{} does not use any information about the structure of the manifold from which the data is drawn. This hurts \Nystrom{} in situations where the manifold is known (\eg a 3-D mesh), since the manifold can be exploited efficiently. When the manifold is unknown, the steps taken to recover it can mean that BHA and other manifold-based methods take longer to set up.

The \Nystrom{} approximation requires \On{nl + l^2} space for storing the matrices \mat{C} and $\mat{W}^{\dagger}$. The runtime is \On{lf} for the computation of \mat{C}, where $f$ is the complexity of evaluating one row of the distance matrix \mat{K}, and \On{l^3} for computing the pseudoinverse of \mat{W}.

\subsubsection{FMDS}
\label{sec:fmds}
In fast multidimensional scaling (FMDS) \cite{FMDS} the distance matrix is approximated as the symmetrized product of an interpolation matrix, $\mat{H}\in \real{n \times l}$, and a matrix $\mat{C}\in \real{l \times n}$ formed by $l$ rows from \mat{K} (similar to \Nystrom{}),
\begin{equation}
  \matamet{K}{\mathit{FMDS}} = \frac{1}{2} (\mat{H} \mat{C} + \mat{C}^\top \mat{H}^\top).
\end{equation}
The interpolation operator \mat{H} is similar but not identical to the BHA interpolation operator \mat{P}. The difference lies in the fact that \mat{P} does exact interpolation, meaning that the values at known points (the landmarks \ve{b}) must be exactly equal to known values. In contrast, \mat{H} allows some small error at the known points, where the amount of error is controlled by a scalar coefficient $\mu$ \cite{FMDS}. It is possible to linearly transform \mat{P} into \mat{H}: $\mat{H} = \mat{P} (\mat{M}_{bb} + \mu \mat{I}_l + \mat{M}_{bu} \mat{P}_u)^{-1} \mu$,
where $\mat{P}_u$ is defined as in Equation \ref{eq:PDefinition}.

It is important to note that FMDS stores much more of the distance matrix in memory than sBHA. Since it uses such a similar method for interpolation, it is expected that FMDS will yield better performance with the same number of landmarks, albeit using much more memory.
Also, the symmetrization step required by FMDS can be expensive for large meshes, whereas SMDS and sBHA are fundamentally symmetric.

\begin{figure*}
	\begin{center}
		\includegraphics[trim=8 0 42 25, clip, width=0.89\columnwidth]{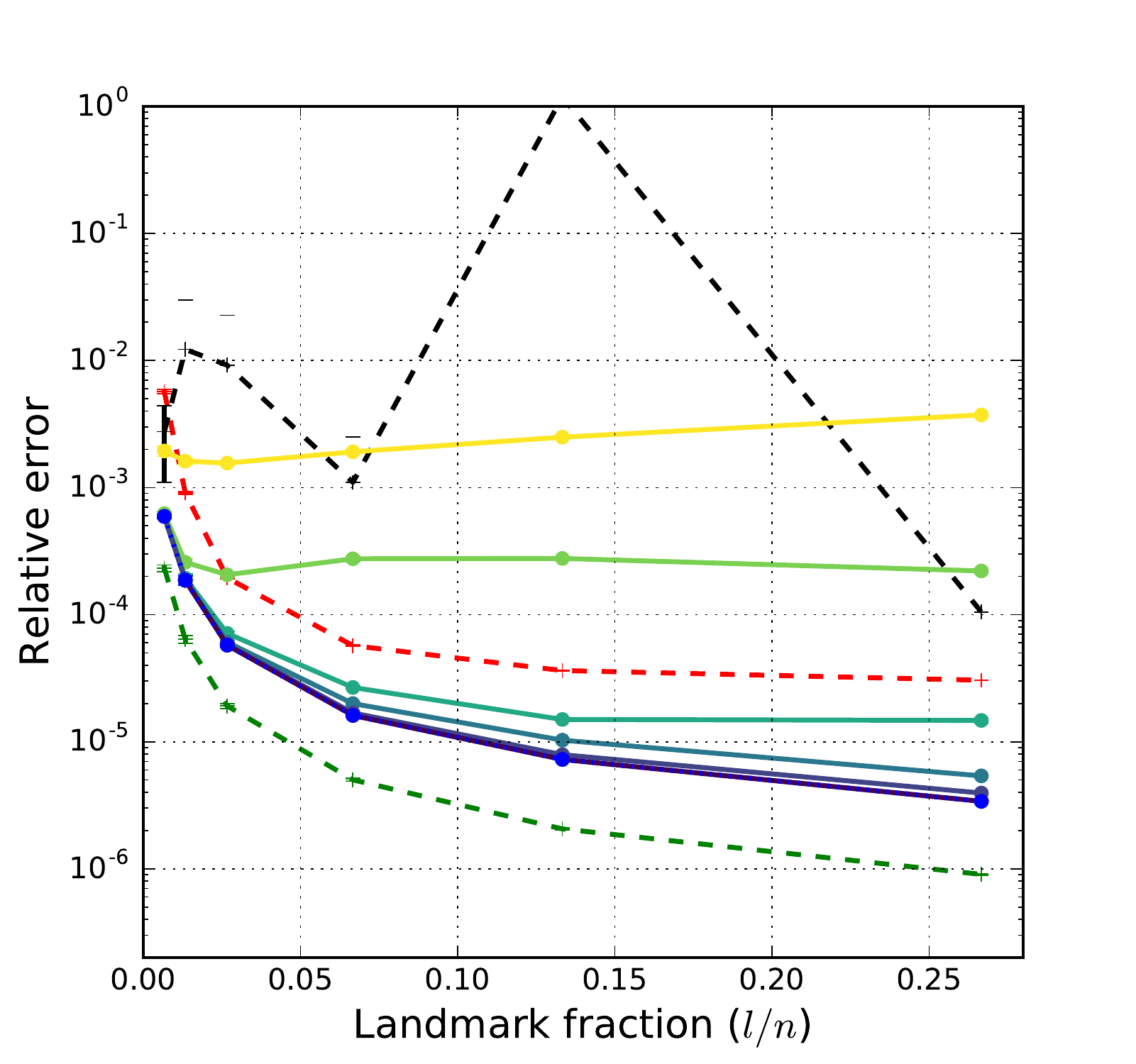}
		\includegraphics[trim=43 0 17 25, clip, width=1.11\columnwidth]{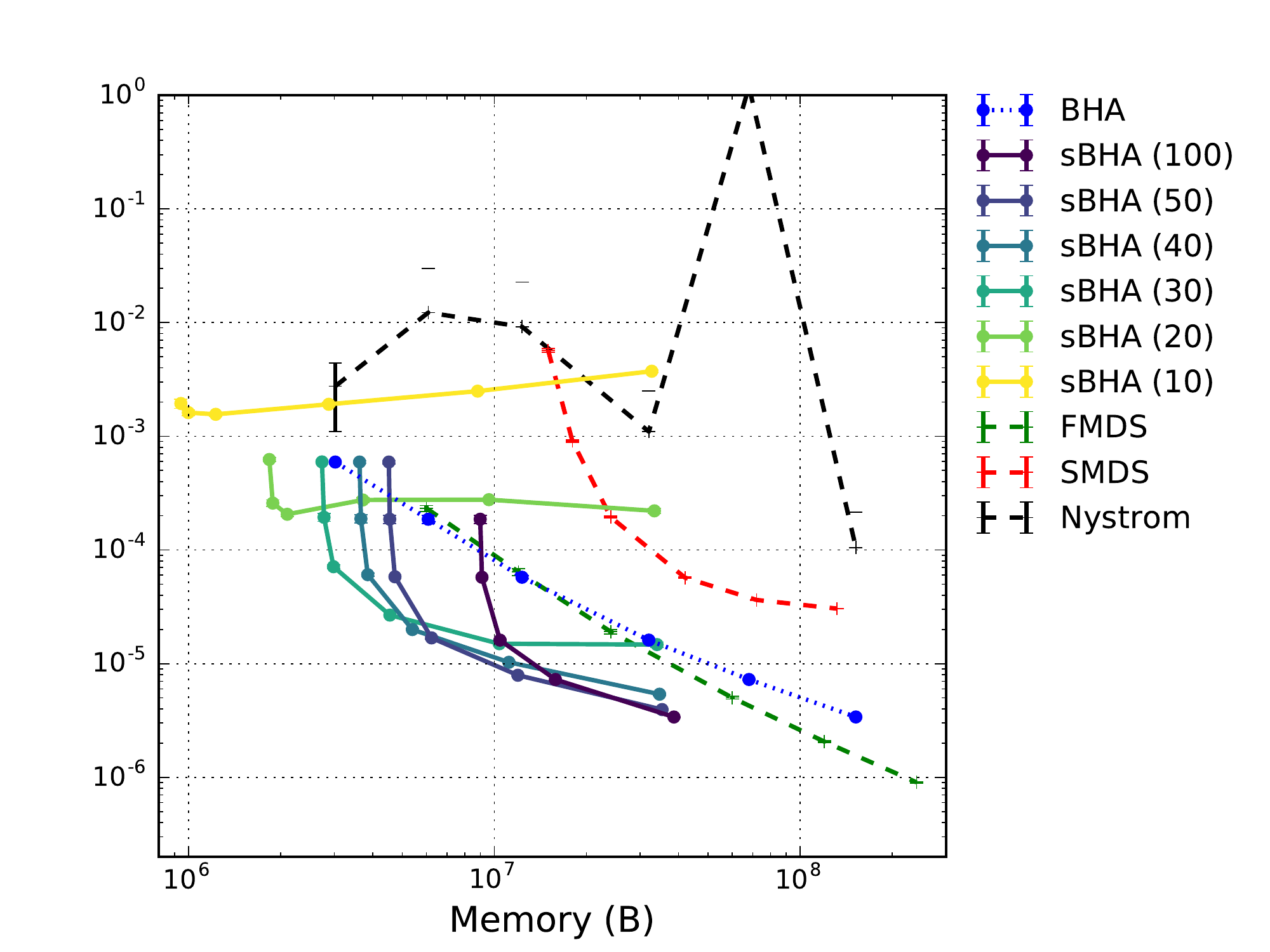}
		\llap{\raisebox{0.5cm}{
			\includegraphics[trim=0 0 160 0, clip, width=2.6cm]{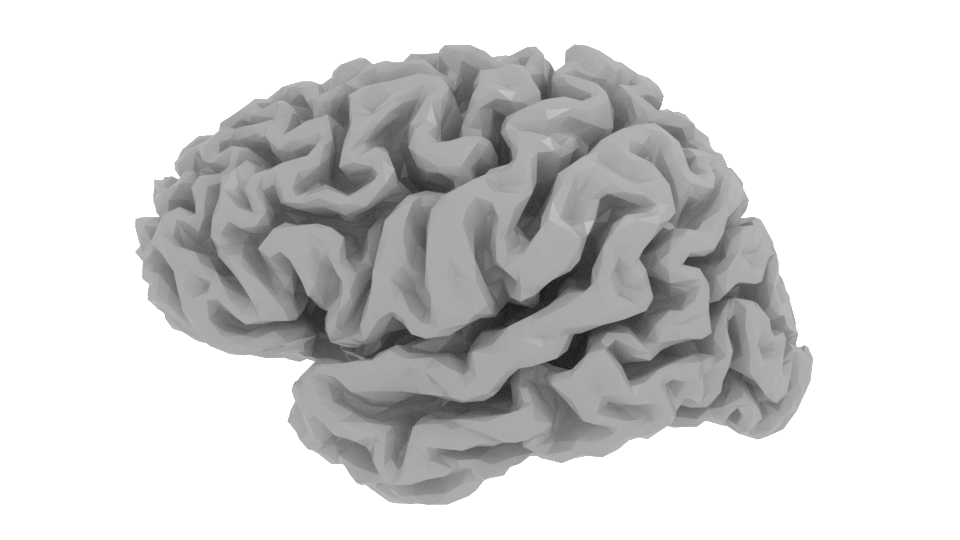}
		}\llap{\raisebox{2.4cm}{{\small \dataset{Brain} \hspace*{0.6cm}}}}}
		\caption{Geodesic distance approximation error for BHA and sBHA with $p_{\mathit{row}}=10\dots100$ (our methods) vs. FMDS, SMDS, and \Nystrom{} on \dataset{Brain} with numbers of landmarks $l$ ranging from 1\% to 25\% of the total number of points $n=7,502$. Each experiment was repeated 10x. \textbf{(Left)} Error vs. Landmark fraction $\nicefrac{l}{n}$. FMDS has the lowest error for each number of landmarks, while \Nystrom{} suffers from numerical instability. There is little difference between BHA and sparse sBHA with $p_{\mathit{row}}=100$ or $p_{\mathit{row}}=50$, but sparser solutions suffer. \textbf{(Right)} Error vs. size of the approximation in memory (bytes). sBHA strongly outperforms FMDS, using 3-4x less memory to achieve the same error rate.}
		\label{fig:sparsity}
	\end{center}
\end{figure*}

\subsubsection{SMDS}
\label{sec:smds}
In spectral multidimensional scaling (SMDS) \cite{SMDS} the distance matrix is also approximated using interpolation, but the dimensionality of the problem is reduced by working in the spectral domain formed by the eigenspace of the LBO. First, sparse eigenvalue decomposition is used to extract the first $m$ eigenvectors $\mat{\Phi}\in\real{n\times m}$ and eigenvalues \mat{\Lambda} of the LBO. Then $l$ landmarks are selected and a smooth interpolation operator $\mat{H}\in \real{m\times l}$ is computed. Finally distances are computed among the landmarks and stored in the symmetric matrix $\mat{W}\in \real{l\times l}$ as in BHA. The approximation is then given as:
\begin{equation}
\matamet{K}{\mathit{SMDS}} = \mat{\Phi} \mat{H} \mat{W} \mat{H}^\top \mat{\Phi}^\top.
\end{equation}
SMDS has several advantages: the resulting matrix is always symmetric, and working in the eigenspace of the LBO reduces dimensionality significantly. However, computing the eigenvectors is extremely costly in runtime, rendering this method generally less useful than FMDS.

\subsubsection{Other methods}
Other related methods include the constant time geodesic (CTG) approximation \cite{Xin} and SpectroMeter (SM) method \cite{Spectrometer}. CTG uses a geometric approach to ``unfold" landmark distances to the entire surface, and thus requires only 3 dimensions but also involves a nonlinear operation (taking the minimum across possible paths). SM uses a spectral decomposition of the Laplace-Beltrami operator (LBO) to rapidly approximate operations used in the heat method for computing geodesics \cite{crane2013}, and to interpolate distances, similarly to SMDS. 


\section{Experimental results}
\label{sec:experiments}
We empirically evaluate BHA and sBHA, and compare to \Nystrom{} \cite{nystrom1930}, FMDS \cite{FMDS}, SMDS \cite{SMDS}, and LMDS \cite{LMDS} in terms of matrix approximation accuracy, MDS stress, memory usage and runtime. We used four point sets: \dataset{Brain} is a 3-D mesh reconstruction of one human cortical hemisphere with $n=7,502$ vertices. \dataset{Bunny} is a 3-D mesh with $n=14,290$ (Stanford Computer Graphics Laboratory). \dataset{Dragon} is a 3-D mesh with $n=1,804,693$ (Stanford Computer Graphics Laboratory). Using quadric decimation in meshlab \cite{meshlab} we created 8 downsampled \dataset{Dragon} meshes with $n=5,000\dots 750,000$. \dataset{TOSCA} \cite{TOSCA} is a set of 148 3-D meshes from 12 categories. Within each category the same mesh appears in different poses.

All experiments were run on a system with two Intel Xeon E5-2699v4 processors (44 cores in total) with 128~GB of physical memory and running Linux. Code\footnote{\url{http://github.com/alexhuth/BHA}} was implemented in Python with optimized and parallelized NumPy and SciPy modules.
BHA, sBHA, FMDS, and SMDS were implemented with sparse Cholesky decomposition from Scikit-Sparse. The FMDS smoothing parameter was set to $\mu=50$ as recommended in \cite{FMDS}. For SMDS $m=200$ eigenvectors were used, as recommended in \cite{SMDS}. Geodesic distances were computed using the approximate geodesics in heat method \cite{crane2013} implemented in pycortex \cite{PYCORTEX}. All reported memory usage is the final memory consumption of the distance matrix approximation, not counting memory used for intermediate steps. 


\begin{figure}
	\begin{center}
			\includegraphics[trim=5 0 20 2, clip,width=0.99\columnwidth]{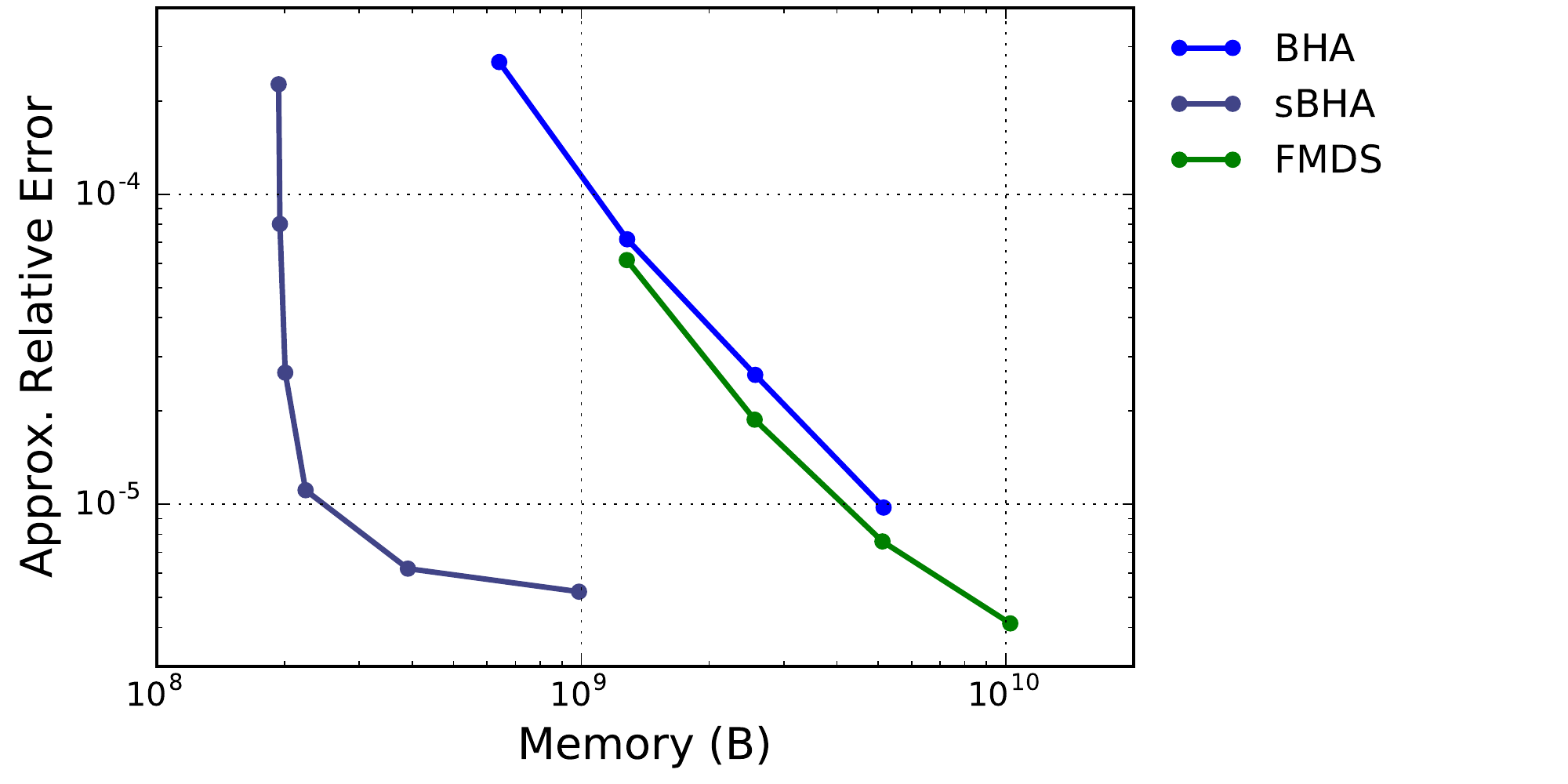}
			\llap{\raisebox{0.45cm}{
    			\includegraphics[trim=0 0 250 0, clip, width=2.2cm]{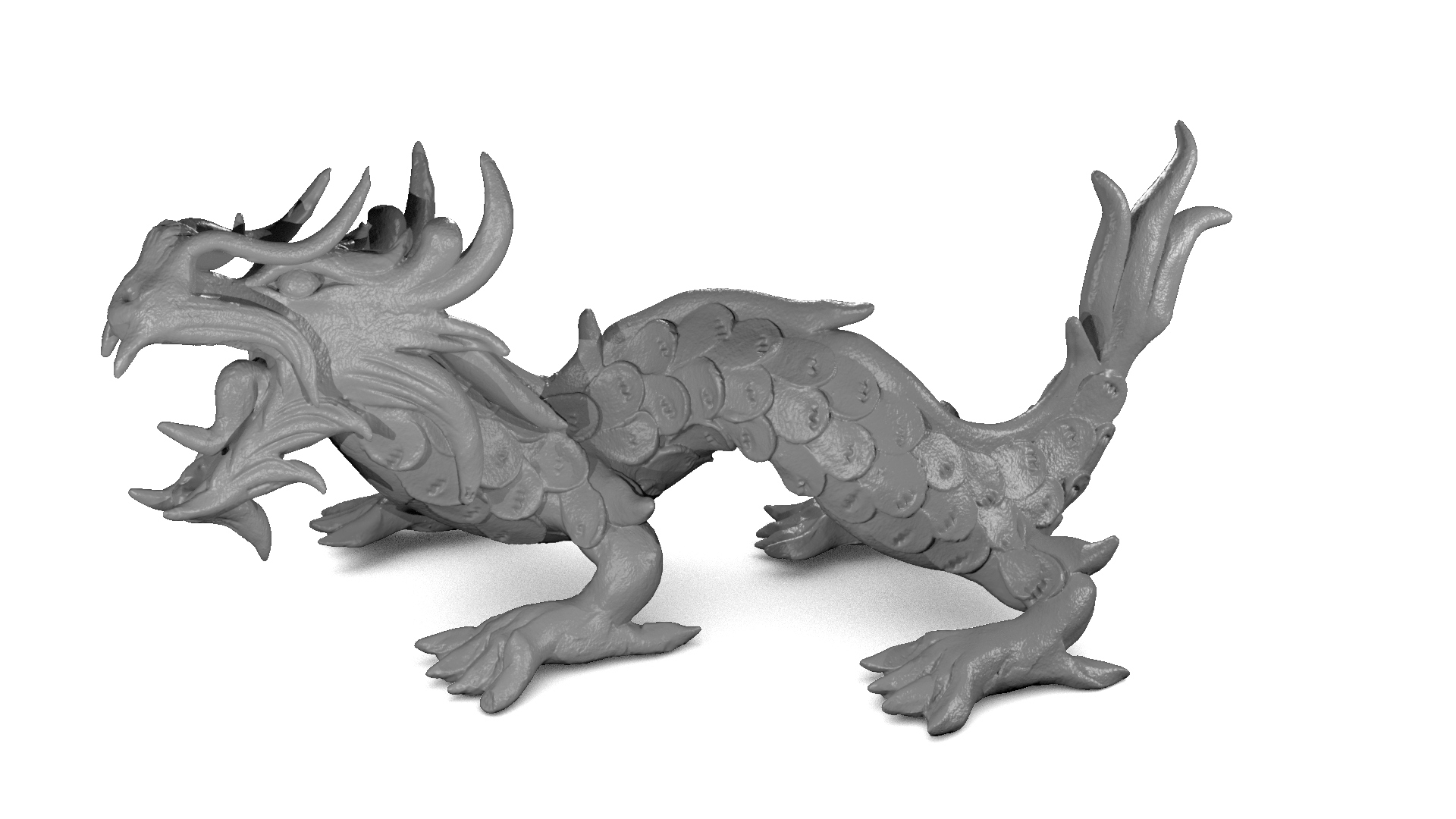}
    		}\llap{\raisebox{2cm}{{\small \dataset{Dragon} \hspace*{0.6cm}}}}}
		\caption{Geodesic distance approximation error for BHA and sBHA with $p_{\mathit{row}}=50$ (our methods) vs. FMDS on \dataset{Dragon320k}, a large mesh with $n=320,003$ vertices, using landmark fractions $\nicefrac{l}{n}=0.0008\dots0.031$. Error is plotted vs. size of the approximation in memory (bytes). sBHA performs extremely well, using about 20x less memory than FMDS to achieve an error of 1 in 100,000.}
		\label{fig:dragon_mem_error}
	\end{center}
\end{figure}

\subsection{Distance matrix approximation}
\label{sec:experiments_sparsity}

Here we evaluate the approximation performance and memory usage of BHA and sBHA versus other methods. We compute low-rank approximations of the geodesic distance matrix for \dataset{Brain} using BHA, sBHA with the number of non-zeros per row $p_{\mathit{row}}$ from 10 to 100, FMDS, SMDS, and \Nystrom{} and then compare to the actual distance matrix. Relative approximation error was measured as $\varepsilon(\mata{K}) = \nicefrac{\normfro{\mata{K}-\mat{K}}^2}{\normfro{\mat{K}}^2}$. With each method we varied the number of landmarks $l$ between roughly 1\% ($l=50$) and 25\% ($l=2,000$) of the total number of points $n$.

Comparing performance as a function of the number of landmarks (Figure \ref{fig:sparsity}, left) shows that FMDS has the lowest error for each value of $l$. BHA performs identically to sBHA with $p_{\mathit{row}}=100$ and is only slightly better than sBHA with $p_{\mathit{row}}=50$. Sparser solutions perform worse. \Nystrom{} performs very badly due to numerical instability of the pseudoinverse $\mat{W}^\dagger$.

However, plotting performance as a function of the size of the approximation in memory (Figure \ref{fig:sparsity}, right) shows that sBHA uses 3-4x less memory to achieve the same level of performance as FMDS. 

The same experiment was performed on \dataset{Bunny} (Supplemental Figures) with similar results for BHA, sBHA, and \Nystrom{}. However, FMDS and SMDS with default parameter settings performed much worse on \dataset{Bunny}, both failing to beat BHA for any given number of landmarks. This suggests that FMDS and SMDS may be more sensitive to hyperparameters than (s)BHA.

To test on a larger problem we also ran BHA, sBHA, and FMDS on \dataset{Dragon320k}, a mesh with $n=320,003$. The full distance matrix was too large to fit in memory, so error was estimated using a random subset of 3000 rows. 
Figure \ref{fig:dragon_mem_error} shows that sBHA strongly outperforms FMDS on this large mesh. To achieve an error of one part in 100,000, sBHA uses roughly 20x less memory than FMDS (200 MB vs. 4 GB). This factor is much larger than for \dataset{Brain}, suggesting that the efficiency gains of sBHA over FMDS grow with the size of the problem. 

\begin{figure}
	\begin{center}
		\includegraphics[trim=0 0 0 0, clip, width=\columnwidth]{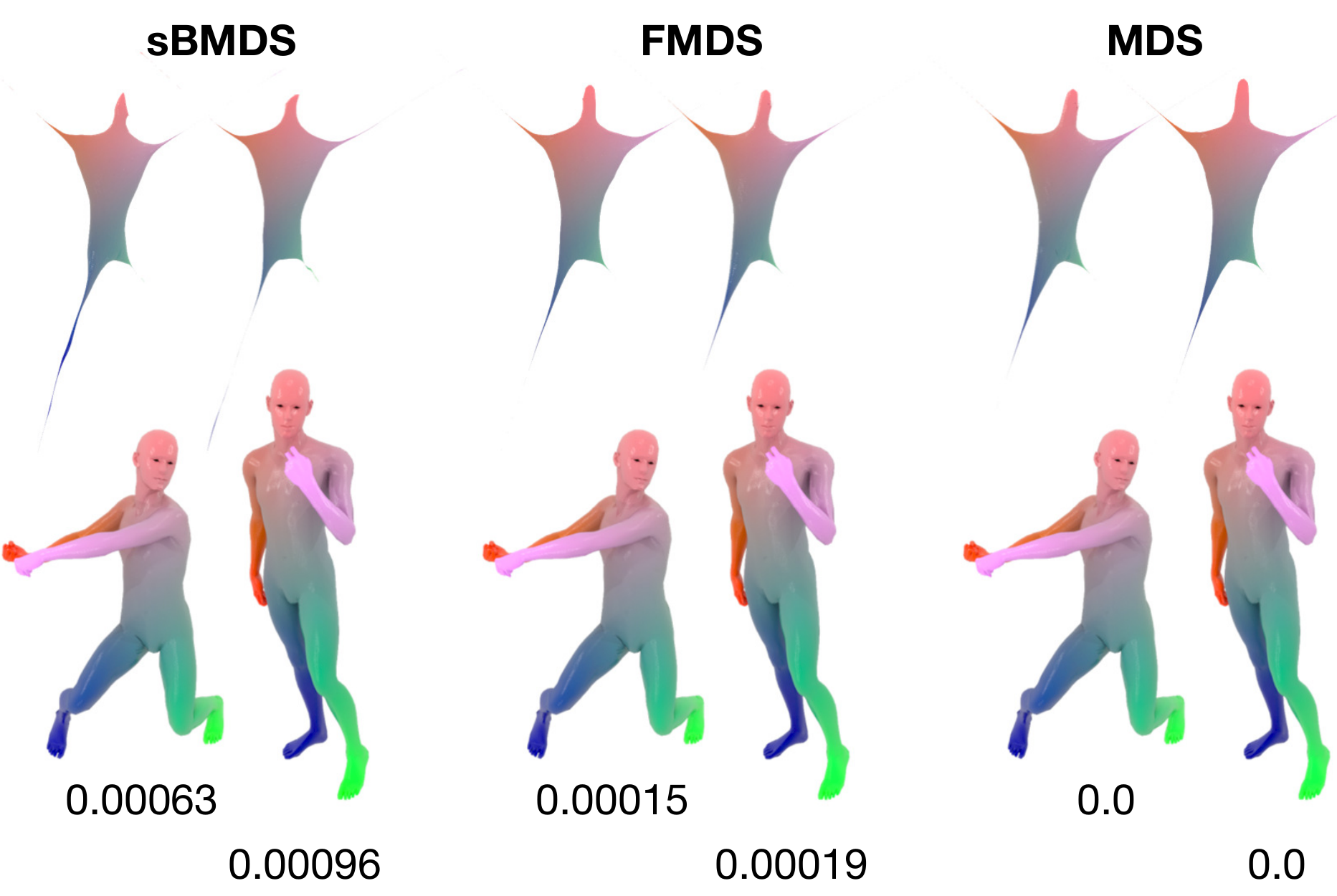}
		\caption{\textbf{(Top)} Canonical forms obtained using sBMDS (our method; using $l=100$ landmarks and $p_{\mathit{row}}=50$), FMDS (using $l=100$ landmarks), and MDS on \dataset{david-1} and \dataset{david-2} from \dataset{TOSCA} and then aligned using ICP. \textbf{(Middle)} Canonical coordinates are mapped to RGB colors on the original meshes. Parts have the same coordinates (and color) despite pose differences. sBMDS yields nearly identical solutions while using much less memory and time than the other methods. \textbf{(Bottom)} Final stress of each approximate MDS solution $-$ stress of the exact solution.}
		\label{fig:canonical_forms}
	\end{center}
\end{figure}

\begin{figure*}
	\begin{center}
			\includegraphics[trim=8 0 42 5, clip, width=0.89\columnwidth]{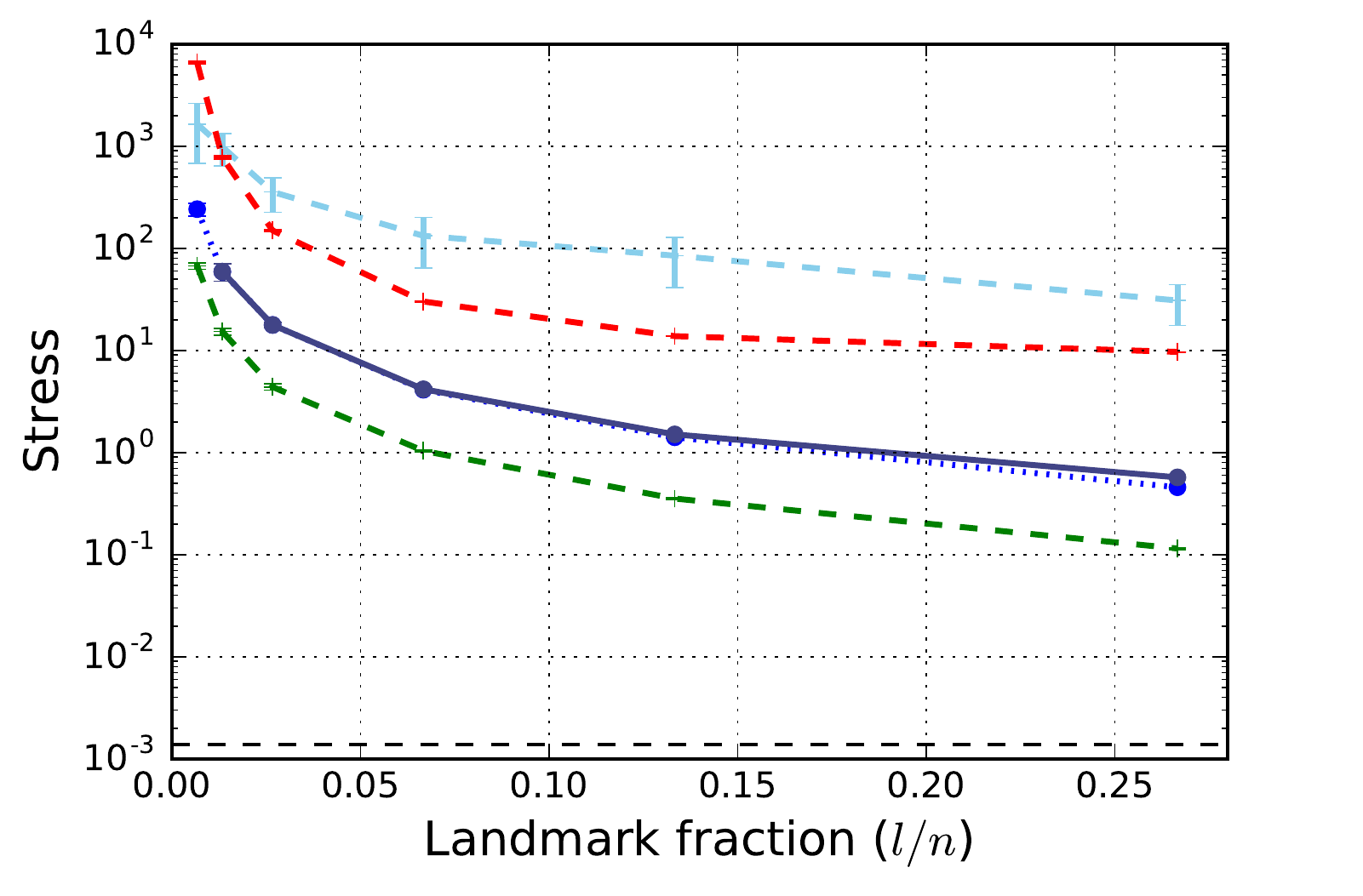}
			\includegraphics[trim=43 0 17 5, clip, width=1.11\columnwidth]{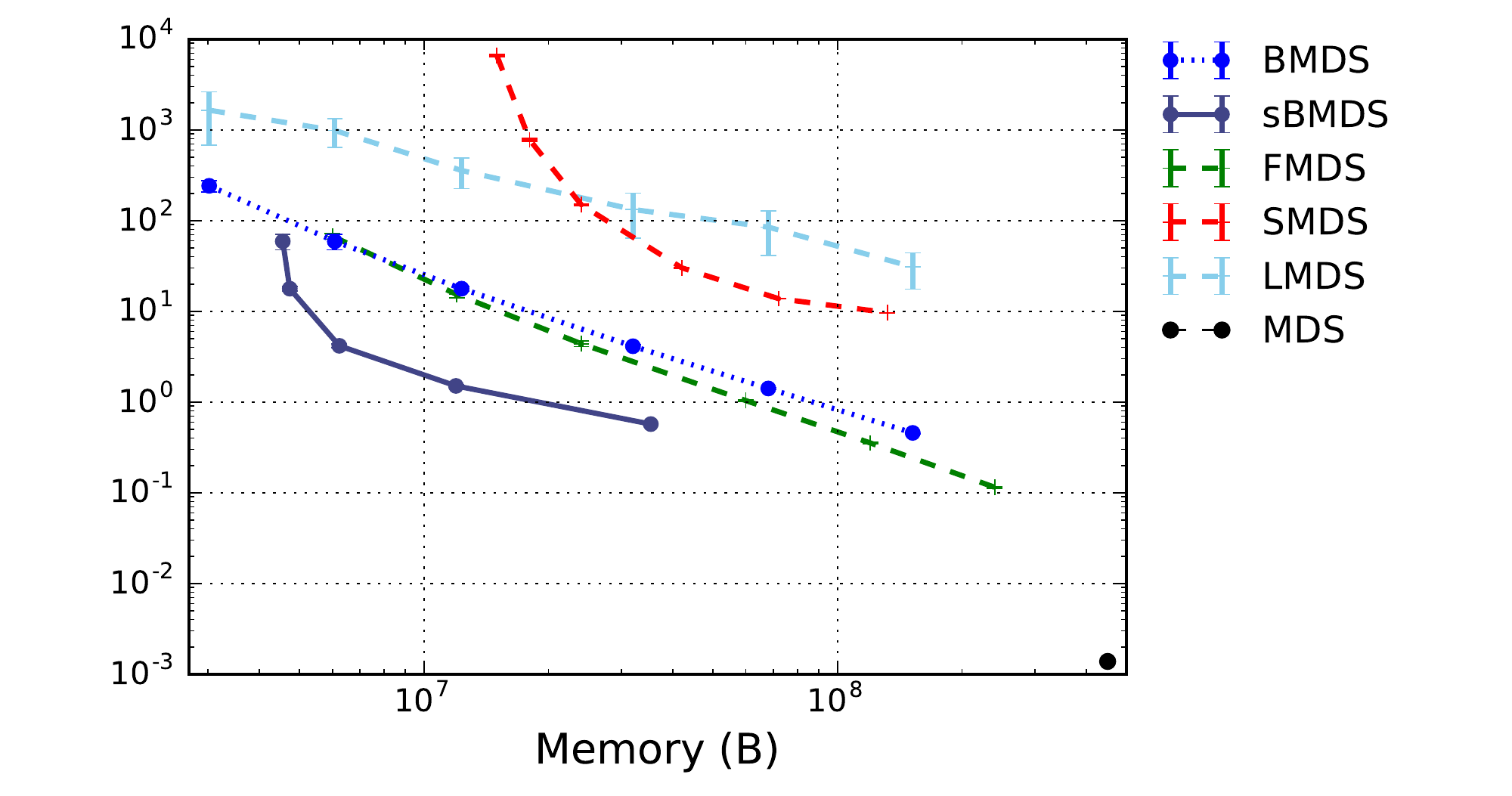}
			\llap{\raisebox{0.5cm}{
			\includegraphics[trim=0 0 160 0, clip, width=2.6cm]{brain7k}
		}\llap{\raisebox{2.4cm}{{\small \dataset{Brain} \hspace*{0.6cm}}}}}
		\caption{MDS stress for BMDS and sBMDS with $p_{\mathit{row}}=50$ (our methods) versus FMDS, SMDS, LMDS, and normal MDS after embedding the geodesic distance matrix for \dataset{Brain} into 3-D. Numbers of landmarks $l$ ranged from 1\% to 25\% of the total number of points $n=7,502$. Each experiment was repeated 10x. \textbf{(Left)} Stress vs. Landmark fraction $\nicefrac{l}{n}$. FMDS has the lowest stress for each number of landmarks. \textbf{(Right)} Stress vs. size of the approximation in memory. sBHA outperforms other methods, using less memory to attain the same stress.}
		\label{fig:methodstress_brain}
	\end{center}
\end{figure*}

\subsection{Obtaining canonical forms using MDS}
FMDS, SMDS, and LMDS were designed specifically for applying MDS to large problems. Using MDS to embed a geodesic distance matrix into 3 dimensions gives the \emph{canonical form} $\mat{Z}\in\real{n\times 3}$, a representation of the dataset that is largely invariant to nonrigid deformations (Figure \ref{fig:canonical_forms}). Comparing canonical forms can reveal whether two meshes are different poses of the same model, and, if so, which parts match one other.

The quality of an MDS solution can be evaluated by computing stress (Equation \ref{eq:MDS_problem}), which measures how different the global geometries of the original and embedded datasets are. We compare MDS results from BMDS and sBMDS with $p_{\mathit{row}}=50$ to FMDS, SMDS, LMDS, and normal MDS.

We first compare stress after applying MDS to \dataset{Brain} (Figure \ref{fig:methodstress_brain}). As in the matrix approximation experiment, FMDS outperforms all other methods when using the same number of landmarks $l$, but sBMDS is much more efficient, achieving the same stress while using much less memory. LMDS performs the worst for most landmark fractions. Comparisons on \dataset{Bunny} were similar to the matrix approximation results (Supplemental Figures).

\begin{figure*}
	\begin{center}
				\includegraphics[trim=0 0 27 15, clip, width=0.89\columnwidth]{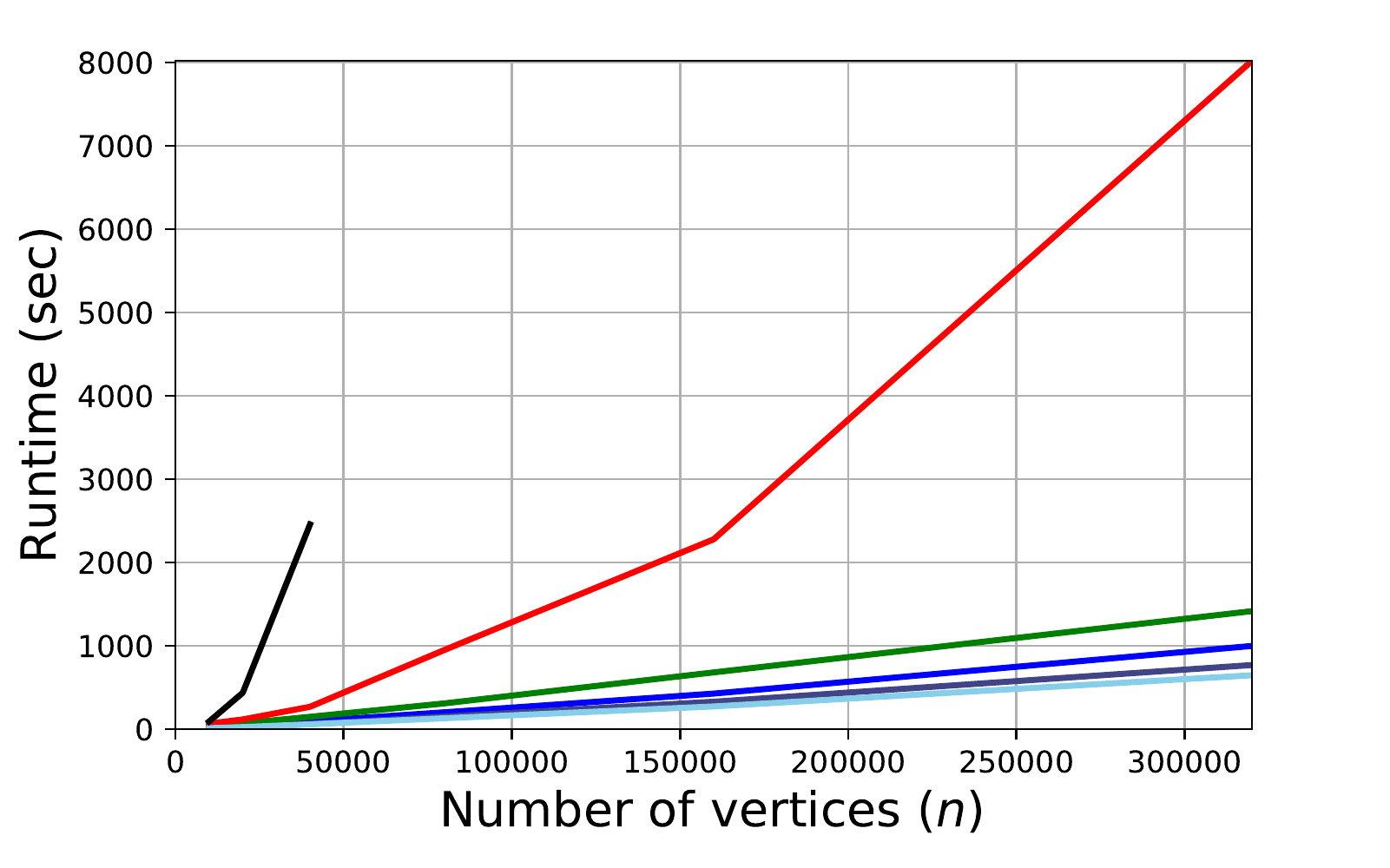}
		\includegraphics[trim=20 0 17 15, clip, width=1.11\columnwidth]{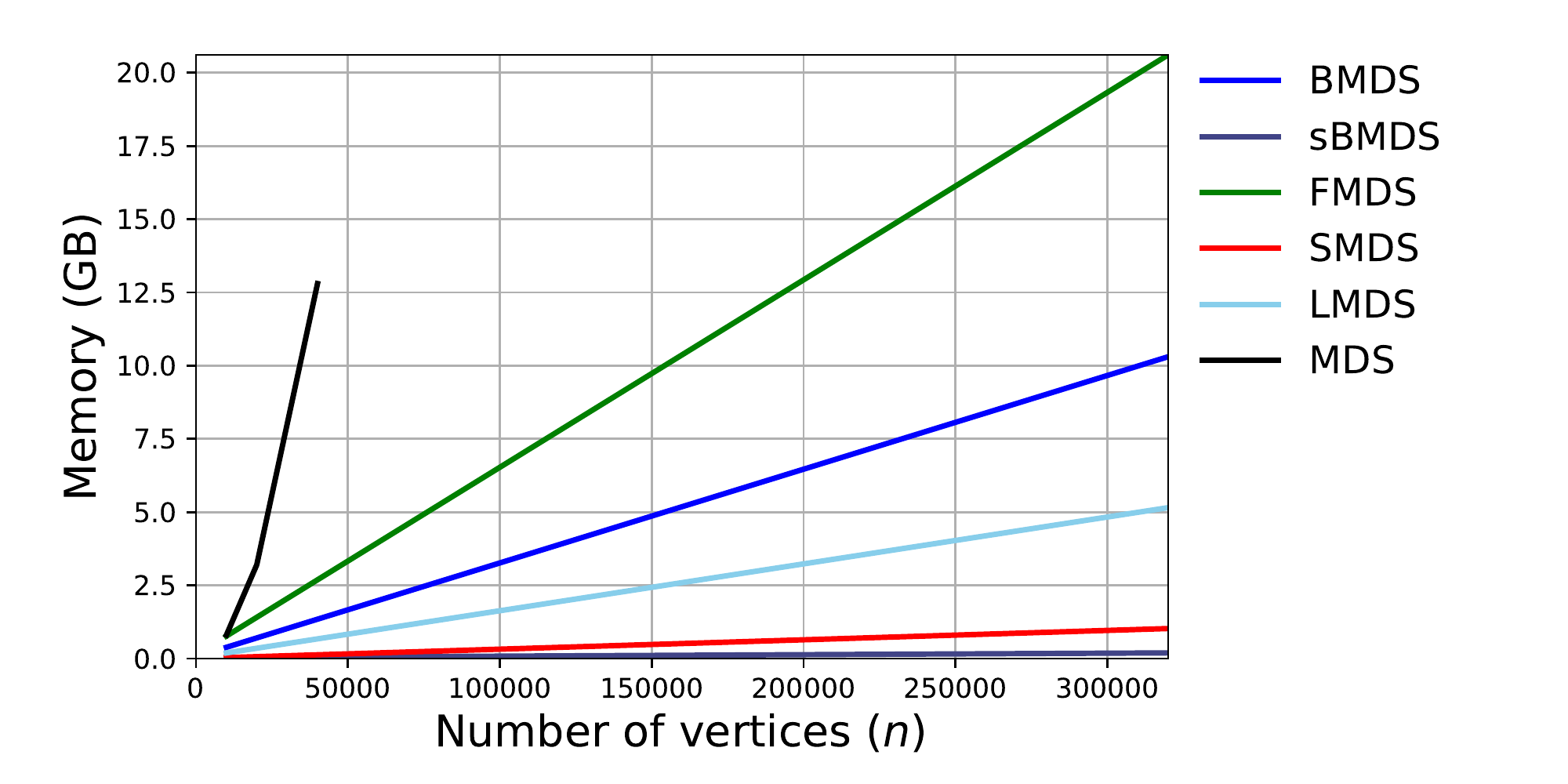}
		\llap{\raisebox{0.35cm}{
				\includegraphics[trim=0 0 320 0, clip, width=2.3cm]{dragon_gray}
			}\llap{\raisebox{1.8cm}{{\small \dataset{Dragon} \hspace*{0.6cm}}}}}
		\caption{Scalability of BMDS and sBMDS with $p_{\mathit{row}}=50$ (our methods) versus other approximations SMDS, FMDS, and LMDS, as well as normal MDS. Each algorithm was used to embed geodesic distance matrices for \dataset{Dragon} meshes with $n=5000\dots 320,000$ vertices into 3 dimensions. \textbf{(Left)} Runtime versus mesh size. \textbf{(Right)} Memory usage versus mesh size.}
		\label{fig:dragon_mds_scaling}
	\end{center}
\end{figure*}

We next compared scalability of the six methods in terms of runtime and memory usage by testing on \dataset{Dragon} meshes having 5,000 to 320,000 vertices. All tests used $l=1,000$ landmarks. Runtime (Figure \ref{fig:dragon_mds_scaling}, left) was best for LMDS, followed closely by sBMDS, BMDS, and FMDS. SMDS was much slower, taking twice as long to embed \dataset{Dragon160k} as sBMDS took to embed the larger \dataset{Dragon320k}. All approximate methods were much faster than normal MDS, which could only be run on meshes up to 40,000 vertices due to memory constraints. Memory usage (Figure \ref{fig:dragon_mds_scaling}, right) scales linearly with mesh size for all methods, but the total memory used varies wildly. sBMDS used by far the least memory, followed by SMDS. FMDS used the most memory.

Finally we studied the quality of the canonical forms obtained using each MDS method. Following \cite{FMDS} we compared 61 meshes coming from 5 different categories (cat, david, horse, lioness, centaur) in \dataset{TOSCA} \cite{TOSCA}. We first obtained a canonical form for each mesh using each method. Examples are shown in Figure \ref{fig:canonical_forms}. Iterated closest point (ICP) was then used to find the distance between each pair of canonical forms, and these distances were submitted to a second stage MDS. The resulting embeddings (Supplemental Figures) clearly cluster according to category for each approximation method.


\subsection{Extremely large mesh}
To demonstrate that our sparse methods, sBHA and sBMDS, can be applied to extremely large problems we used sBMDS to obtain the canonical form for \dataset{Dragon1800k}, a mesh with 1.8 million vertices. Using $l=50,000$ landmarks the approximate geodesic distance matrix is only 20.9 GB, three orders of magnitude smaller than the full 26 TB matrix. The canonical form clearly distinguishes the extremities of the mesh (Figure \ref{fig:dragon_mds}), suggesting that sBMDS was successful at recovering the canonical form. 

\section{Conclusions}
\label{sec:conclusions}
The sBHA method described here offers a sparse alternative to approximation methods like FMDS, SMDS, and \Nystrom{}. Sparsity allows sBHA to be extremely efficient in both memory and evaluation time. Results show that sBHA can be used for the same applications and can yield equally accurate approximations using 20x less memory than other methods. The greatest value of sBHA is for scaling to very large problems where the accuracy of current methods is limited by memory.

One key improvement to (s)BHA would be a way to automatically select the number of landmarks that gives an efficient but accurate approximation. Another way forward could be to split the difference between FMDS and sBHA by saving more of the precomputed distances than sBHA does but fewer than FMDS. Ultimately it will also be important to study the theoretical properties of this method and provide bounds on approximation error. Nevertheless, these results show that sBHA can be extremely beneficial in some settings, and is immediately applicable.

{\small
\bibliographystyle{ieee}
\bibliography{bibliography}
}

\end{document}